\documentclass{article}

\usepackage{microtype}
\usepackage{graphicx}
\usepackage{subfigure}
\usepackage{booktabs} 

\usepackage{hyperref}

\usepackage[accepted]{icml2025}

\usepackage{amsmath}
\usepackage{amssymb}
\usepackage{mathtools}
\usepackage{amsthm}

\usepackage[capitalize,noabbrev]{cleveref}

\usepackage{enumitem}   

\theoremstyle{plain}
\newtheorem{theorem}{Theorem}[section]

\newtheorem{corollary}[theorem]{Corollary}
\theoremstyle{definition}
\newtheorem{definition}[theorem]{Definition}

\theoremstyle{remark}

\usepackage[textsize=tiny]{todonotes}

\icmltitlerunning{General and Estimable Learning Bound Unifying Covariate and Concept Shifts}

\begin{document}

\twocolumn[
\icmltitle{General and Estimable Learning Bound Unifying Covariate and Concept Shifts}



\icmlsetsymbol{equal}{*}

\begin{icmlauthorlist}
\icmlauthor{Hongbo Chen}{yyy}
\icmlauthor{Li Charlie Xia}{yyy}
\end{icmlauthorlist}

\icmlaffiliation{yyy}{Department of Statistics and Financial Mathematics, South China University of Technology, Guangzhou, China}

\icmlcorrespondingauthor{Hongbo Chen}{hongboc616@gmail.com}
\icmlcorrespondingauthor{Li Charlie Xia}{lxia@scut.edu.cn}

\icmlkeywords{Machine Learning, ICML}

\vskip 0.3in
]



\printAffiliationsAndNotice{}  

\begin{abstract} Generalization under distribution shift remains a core challenge in modern machine learning, yet existing learning bound theory is limited to narrow, idealized settings and is non-estimable from samples. In this paper, we bridge the gap between theory and practical applications. We first show that existing bounds become loose and non-estimable because their concept shift definition breaks when the source and target supports mismatch. Leveraging entropic optimal transport, we propose new support-agnostic definitions for covariate and concept shifts, and derive a novel unified error bound that applies to broad loss functions, label spaces, and stochastic labeling. We further develop estimators for these shifts with concentration guarantees, and the DataShifts algorithm, which can quantify distribution shifts and estimate the error bound in most applications - a rigorous and general tool for analyzing learning error under distribution shift.
\end{abstract}

\section{Introduction}

With the growth of data and computing power, supervised learning has achieved remarkable success. Nonetheless, traditional supervised learning assumes that the training data (source domain) and the test or deployment data (target domain) share the same distribution. However, in many real-world applications, the test data distribution can differ substantially from the training distribution, and this discrepancy can significantly impact model performance in the target domain. To analyze such challenges, researchers theorized that the distributions between the source and target domains are shifted and developed methods to assess how learners trained in the source domain could perform on the target domain. Depending on whether the target domain data is accessible, the problems are further categorized into domain adaptation and domain generalization.

Recent years have seen an abundance of theoretical results \citep{ben2006analysis,ben2010theory,zhao2019learning,ye2021towards} and algorithms \citep{sugiyama2007covariate,glorot2011domain,sun2016deep,ganin2016domain,pei2018multi,arjovsky2019invariant,krueger2021out,rame2022fishr} in these areas. Theoretical results on distribution shift usually bound a model's target domain error by its source domain error plus a measure of the distribution shift. The shift is further dissected into X (covariate) and Y$\mid$X (concept) shifts \citep{sugiyama2007covariate,moreno2012unifying,liu2021towards,cai2023diagnosing,zhang2023nico++}. Two early studies \citep{ben2006analysis,ben2010theory} proposed using $\mathcal{H}$-divergence to measure  X shift and derived an error bound for binary classification. Later works proposed measuring  X shift using the maximum mean discrepancy (MMD) \citep{long2015learning} and the Wasserstein distance \citep{shen2018wasserstein,courty2017joint}. \citet{zhang2020unsupervised} and \citet{zhang2019bridging} proposed more complex metrics for the X shift and obtained bounds for multiclass classification. \citet{chen2023algorithm} gave a bound in a multi-source setting. In general, these results focused only on X shift and additionally relied on a joint-error term between the source and target domains. Lately, \citet{zhao2019learning} pointed out that this cross-domain joint-error term is loose, and alternatively obtained a bound that explicitly considers both X shift and Y$\mid$X shift for binary classification, and \citet{zhang2023nico++} extended the theory to multi-class classification.

In this paper, we focus on two key problems that still block the practical use of existing theories for distribution shift: 

\begin{itemize}[leftmargin=0.6cm]  
  \item \textbf{Estimability.} Although existing bounds provide a preliminary
        definition of Y$\mid$X shift, this definition—and the accompanying error
        bound—are \emph{not} estimable. As a result, one cannot rigorously quantify
        Y$\mid$X shift on real data, nor assess its impact on model performance.

  \item \textbf{Generalizability.} Existing bounds rely on restrictive assumptions:
        they require deterministic labeling, omitting label noise and latent
        confounders commonly seen in practice, and they only apply to classification with absolute error, excluding regression tasks and broader loss families.
\end{itemize}

We aim to provide a more general error bound on distribution shift that widely applies to stochastic labeling, most supervised tasks, and general loss functions. Moreover, one shall be able to accurately estimate it from real-world samples, thereby offering a rigorous tool for quantifying and analyzing distribution shifts to the community.

Specifically, our key observation is that if X shift occurs, the supports of the source and target covariate distributions may not overlap. Such a support mismatch renders the existing theories' Y$\mid$X shift ill-defined, fundamentally causing their error bounds non-estimable and loose. Our key innovation is to employ the entropic optimal transport to redefine X shift, and introduce two new types of Y$\mid$X shift: total point Y$\mid$X shift and total pair Y$\mid$X shift. The total point Y$\mid$X shift extends the existing Y$\mid$X shift to the stochastic labeling and general label space; however, it still remains ill-defined when support mismatch. In contrast, the total pair Y$\mid$X shift we further propose, which depends on the optimal transport plan of X shift, stays well-defined even when supports mismatch. Based on that, we derived a new error bound that considers both the X and the total pair Y$\mid$X shifts. Our new bound naturally generalizes to binary, multi-class, and regression tasks, relying only on the Lipschitz continuity of the hypothesis $h$ and loss $\ell$, and is agnostic to specific hypothesis space, label space, or loss function.

Since our total pair Y$\mid$X shift is well-defined regardless of support mismatch, X and Y$\mid$X shifts, and our new error bound becomes estimable from samples. Nonetheless, for X shift, the traditional plug-in estimator for entropic optimal transport tends to overestimate due to the curse of dimensionality. We thus further developed a debiased estimator that remains accurate in high dimensions. For our total pair Y$\mid$X shift, we also proposed an estimator. We proved both estimators' concentration inequalities to their true values. By combining the estimators with existing methods for computing Lipschitz constants of learners, we developed the DataShifts algorithm, which can quantify X and Y$\mid$X shifts and estimate our new error bound from general labeled data.

\textbf{Contributions.} In summary, our major contributions are:
\begin{itemize}[leftmargin=0.6cm]
\item We show that support mismatch makes the existing Y$\mid$X shift ill-defined, leading to loose and non-estimable error bounds. We propose a well-defined concept shift (total pair Y$\mid$X shift) and a new error bound based on it. Our bound covers most practical learning scenarios and is estimable from real data.
\item For the X and total pair Y$\mid$X shifts in our bound, we propose new estimators and prove their concentration inequalities. On this basis, we introduce the DataShifts algorithm, which can quantify X and Y$\mid$X shifts in real data and estimate our error bound, supplying a rigorous and general tool for measuring distribution shift and analyzing learning performance.
\end{itemize}

The rest of the paper is organized as follows: Section~\ref{sec:preliminaries} is the preliminary. In Section~\ref{sec:theory}, we give our theoretical results, including new definitions of X and Y$\mid$X shifts and our learning bound. Section~\ref{sec:estimation} focuses on the estimation of the bound, including two new estimators and their concentration inequalities, as well as the DataShifts algorithm. We conduct experiments and validate our theory in Section~\ref{sec:experiments}.


\section{Preliminary}
\label{sec:preliminaries}

\subsection{Problem Setup}

Our problem is to bound the error of models under distribution shift. Let $\mathcal{X}$ and $\mathcal{Y}$ be the covariate space and the label space, respectively. Let $\mathcal D_{XY}^{S}$ and $\mathcal D_{XY}^{T}$ be the joint distributions of covariates and labels on $\mathcal X\times\mathcal Y$ for the source and target domains, respectively. Let $\mathcal D_{X}^{S}$, $\mathcal D_{X}^{T}$ be their covariate marginals on $\mathcal X$. And for any $x\in\mathcal X$, we let $\mathcal D_{Y|X=x}^{S}$ and $\mathcal D_{Y|X=x}^{T}$ be the conditional label distributions at $x$ in the source and target domain, respectively. Let $\mathcal Y^{\prime}$ be the output space of the learner, $\ell:\mathcal Y\times\mathcal Y^{\prime}\to\mathbb R$ the loss, and $\mathcal H\subseteq\{g:\mathcal X\to\mathcal Y^{\prime}\}$ the hypothesis space. For a hypothesis $h\in\mathcal H$, the learning errors for source and target domains are defined as: 
\[
\epsilon_S(h) = \mathbb E_{(x_S,y_S)\sim\mathcal D_{XY}^{S}}\bigl[\ell(y_S,h(x_S))\bigr] \\
\]
\[
\epsilon_T(h) = \mathbb E_{(x_T,y_T)\sim\mathcal D_{XY}^{T}}\bigl[\ell(y_T,h(x_T))\bigr]
\]
We hope to bound $\epsilon_T(h)$ by $\epsilon_S(h)$ and a measure of distribution shift, consistent with existing theoretical results \citep{ben2006analysis,zhao2019learning}. Notably, the space $\mathcal X$ can be the raw feature space or a representation space output by an upstream learner \citep{ben2006analysis}. Our theory treats them in the same way, so the results apply to both raw data and learned representations.

\paragraph{Stochastic and deterministic labeling.} Above, we assume that the label $y$ follows the conditional distribution at point $x$: $y\sim\mathcal D_{Y|X=x}$; this is the stochastic labeling setting \citep{zhao2019learning}. It enables our theory to accommodate latent confounders and label noise, which are common in practice. In contrast, existing theories oversimplify by using deterministic labeling with labeling function $f:\mathcal X\to\mathcal Y$ and $y=f(x)$, which is a special case of stochastic labeling, when the conditional distribution collapses to a Dirac mass at $f(x)$: $\mathcal D_{Y|X=x}=\delta_{f(x)}$.

\subsection{Support Mismatch and Ill-Defined Y$\mid$X Shift}

We first demonstrate that support mismatch leads to an ill-defined Y$\mid$X shift, a key flaw in existing theory.

\begin{definition}[Support]
\label{def:support}
Let $\mu$ be a probability measure on the topological space $(\mathcal X,\tau)$. Its support is defined as:
\[
\operatorname{supp}(\mu) \;=\; \{\,x\in\mathcal X \mid \forall\,U\in\tau,\;x\in U,\;\mu(U)>0\}
\]
\end{definition}
$\operatorname{supp}(\mu)$ is the closure of every region where $\mu$ has positive measure, or equivalently, the complement of the union of all $\mu$-null open sets:
\[
\operatorname{supp}(\mu) \;=\; \mathcal X \setminus \Bigl\{\,\bigcup \{\,U\in\tau \mid \mu(U)=0\}\Bigr\}
\]

\begin{definition}[Support Mismatch]
\label{def:support-mismatch}
For two probability measures $P_X,P_X^{\prime}$ on $(\mathcal X,\tau)$, if $\operatorname{supp}(P_X^{\prime}) \setminus \operatorname{supp}(P_X) \;\neq\; \varnothing$, then their supports mismatch.
\end{definition}

\begin{corollary}
\label{cor:support-mismatch-null-set}
If $\operatorname{supp}(P_X^{\prime}) \setminus \operatorname{supp}(P_X) \neq \varnothing$, then there exists a $P_X$-null set $Z\in\mathcal B_{\mathcal X}$ such that $P_X(Z)=0$ and $P_X^{\prime}(Z)>0$, where $\mathcal B_{\mathcal X}$ is the Borel $\sigma$-field on $\mathcal X$.
\end{corollary}

Support mismatch commonly breaks the rigor of theoretical analyses. Some studies explicitly exclude it from analysis by extra assumptions, such as done in the absolute continuity in measure theory \citep{duncan1970absolute}, the positivity assumption in causal inference \citep{cole2009consistency}, or support overlap in importance sampling \citep{gelman1998simulating}.

\begin{definition}[Conditional Probability]
\label{def:conditional-probability}
Let $(\Omega,\mathcal F,P)$ be a probability space and $X:\Omega\to\mathcal X$ a random variable with distribution $P_X$. For $A\in\mathcal F$, a $P_X$-measurable function $P(A\mid X=x):\mathcal X\to[0,1]$ is the conditional probability of $A$ given $X$ if:
\[
\forall B\in\mathcal B_{\mathcal X},\,
P(A\cap\{X\in B\}) = \int_B P(A\mid X=x)\,dP_X(x)
\]
\end{definition}
The conditional probability is unique almost everywhere with respect to $P_X$ (unique $P_X$-a.e.) \citep{kallenberg1997foundations}. That is, for any $P_X$-null set $Z$ with $P_X(Z)=0$, $\int_{Z} P(A\mid X=x)\,dP_X(x) = 0$, 
implying that $P(A\mid X=x)$ can be assigned arbitrarily on $Z$ without affecting its overall properties. Intuitively, it means discussing conditional probabilities on a null set is meaningless \citep{hajek2003conditional}.

\paragraph{Remark (Ill-Defined Conditional Probability Expectation)} If there exists $Z\in\mathcal B_{\mathcal X}$ with $P_X(Z)=0$ and $P_X^{\prime}(Z)>0$, then the expectation of the conditional probability $P(A\mid X=x)$ with respect to $P_X^{\prime}$:
\begin{align*}
\mathbb E_{P_X^{\prime}}[P(A\mid X=x)] 
= \int_{\mathcal X\setminus Z} P(A\mid X=x)\,dP_X^{\prime}(x)
  \\+ \int_Z P(A\mid X=x)\,dP_X^{\prime}(x)
\end{align*}
is \emph{ill-defined}. 

Because $P_X^{\prime}(Z)>0$ while $P(A\mid X=x)$ is arbitrary on the $P_X$-null set $Z$, the second integral $\int_{Z} P(A\mid X=x)\,dP_{X}^{\prime}(x)$ — and thus $\mathbb E_{P_{X}^{\prime}}[P(A\mid X=x)]$ — is arbitrary. Hence, support mismatch leads to the ill-defined expectation of conditional probability. This issue directly impacts the definition and computation of Y$\mid$X shift in existing theories, since the Y$\mid$X shift is typically formulated as an expectation over conditional labeling distributions or its collapsed form labeling functions (see Remark of Theorem~\ref{thm:existing-bound}). When the expectation of conditional probability itself is ill-defined due to support mismatch, the Y$\mid$X shift becomes ill-defined as well, making it impossible to quantify or estimate rigorously.

\subsection{Existing Learning Bound with Distribution Shift}

With deterministic labeling, \citet{zhao2019learning} used $\mathcal H$-divergence \citep{ben2006analysis} to derive an error bound for soft-label binary classification under X and Y$\mid$X shifts:

\begin{theorem}[Existing Bound]
\label{thm:existing-bound}
Let $\mathcal D_{X}^{S}, \mathcal D_{X}^{T}$ be the covariate distributions and $f_S,f_T:\mathcal X\to[0,1]$ the labeling functions for the source and target domain. Using absolute error loss $|\cdot|$, for any hypothesis space $\mathcal H\subseteq[0,1]^{\mathcal X}$ define:
\[
\tilde{\mathcal H}
= \bigl\{\operatorname{sgn}(|h(x)-h^{\prime}(x)| - t)
\;\bigm|\; h,h^{\prime}\in\mathcal H,\; t\in[0,1]\bigr\},
\]
then for any $h\in\mathcal H$,
\begin{align}
\epsilon_T(h)
&\;\le\;
\epsilon_S(h)
+ d_{\tilde{\mathcal H}}\!\bigl(\mathcal D_{X}^{S},\mathcal D_{X}^{T}\bigr)
+ \min\Bigl\{
    \mathbb{E}_{x \sim \mathcal D_{X}^{S}}
      \bigl[\,|f_S(x)-f_T(x)|\,\bigr], \notag \\[2pt]   
&\qquad
    \mathbb{E}_{x \sim \mathcal D_{X}^{T}}
      \bigl[\,|f_S(x)-f_T(x)|\,\bigr]
  \Bigr\} \label{eq:existing-bound}                     
\end{align}

Here, $d_{\tilde{\mathcal H}}$ measures X shift, and the two expectations $\mathbb E_{x\sim\mathcal D_{X}^{S}}[|f_S(x)-f_T(x)|]$, $\mathbb E_{x\sim\mathcal D_{X}^{T}}[|f_S(x)-f_T(x)|]$ are both Y$\mid$X shift but measured with source or target covariate, with the smaller one taken in the bound.
\end{theorem}

\paragraph{Remark (Limitations of Theorem~\ref{thm:existing-bound})}
First, it is highly specialized, applying only to deterministic labeling, binary classification, and absolute error loss. Moreover, when X shift causes support mismatch : $\operatorname{supp}(\mathcal D_{X}^{T})\setminus\operatorname{supp}(\mathcal D_{X}^{S})\neq\varnothing$, the conditional distribution $\mathcal D_{Y|X=x}^{S}$ (i.e., $f_S(x)$) is arbitrary on the mismatched region, so the Y$\mid$X shift $\mathbb E_{x\sim\mathcal D_{X}^{T}}[|f_S(x)-f_T(x)|]$ is ill-defined. Similarly, if $\operatorname{supp}(\mathcal D_{X}^{S})\setminus\operatorname{supp}(\mathcal D_{X}^{T})\neq\varnothing$, the other Y$\mid$X shift term is also ill-defined. This causes two problems:

\begin{enumerate}[nosep,leftmargin=*]  
  \item \textbf{Loose bound.} The ill-defined $Y|X$ shifts can take arbitrary values,
        making the bound in Eq.~\eqref{eq:existing-bound} loose.
  \item \textbf{Non-estimable.} When $f_S(x)$ is unknown, we cannot sample it outside 
        $\operatorname{supp}(\mathcal D_{X}^{S})$;
        hence, under support mismatch the expectations
        $\mathbb E_{x\sim\mathcal D_X^{T}}\!\bigl[|f_S(x)-f_T(x)|\bigr]$
        is inherently non-estimable, as well as $\mathbb E_{x\sim\mathcal D_X^{S}}\!\bigl[|f_S(x)-f_T(x)|\bigr]$.
\end{enumerate}

\subsection{Entropic Optimal Transport}

We introduce entropic optimal transport, which we will employ to redefine X and Y$\mid$X shifts properly. It measures the distance between two probability distributions, augmenting conventional optimal transport with a relative-entropy regularizer, yielding essential desirable properties.

\begin{definition}[Entropic Optimal Transport]
\label{def:entropic-ot}
Given two probability distributions $\mathbb P$ and $\mathbb Q$ on the metric space $(\Omega,\rho)$ and a parameter $\beta\ge0$, the order-1 entropic optimal transport is defined as:
\begin{align}
W_{\beta}(\mathbb P,\mathbb Q)
&=
\inf_{\gamma\in\Gamma(\mathbb P,\mathbb Q)}
\biggl\{
    \int \rho(x_1,x_2)\,d\gamma(x_1,x_2) \notag \\[4pt]
&\qquad \qquad \,+\beta\,H\bigl(\gamma \bigm| \mathbb P\otimes\mathbb Q\bigr)
\biggr\} \label{eq:entropic-OT}
\end{align}

where $
H\!\bigl(\gamma \bigm| \mathbb P\otimes\mathbb Q\bigr)\!=\!
\int \log\Bigl(\tfrac{d\gamma(x_1,x_2)}{d\mathbb P(x_1)\,d\mathbb Q(x_2)}\Bigr)\,d\gamma(x_1,x_2)$.
Here, $\rho(x_1,x_2)$ is the cost of transporting mass between points, $\Gamma(\mathbb P,\mathbb Q)$ is the set of joint distributions with marginals $\mathbb P$ and $\mathbb Q$, and $\gamma$ is a transport plan. Hence, $W_\beta(\mathbb P,\mathbb Q)$ is the minimum total transport cost between $\mathbb P$ and $\mathbb Q$ under an entropy regularizer. When $\beta=0$, this reduces to the classical Wasserstein-1 distance, denoted by $W_{1}(\mathbb P,\mathbb Q)$. In practice, a small positive $\beta$ is chosen, making $W_{\beta}$ close to $W_{1}$ \citep{carlier2017convergence,carlier2023convergence}.
\end{definition}

\begin{corollary}
\label{cor:entropic-properties}
If $\beta>0$, entropic optimal transport has the following properties:
\begin{itemize}[label={},nosep,leftmargin=*]  
  \item (i) The minimization problem is $\beta$-strongly convex in $\gamma$.
  \item (ii) The optimal transport plan $\gamma^{*}$ is unique.
  \item (iii) $\operatorname{supp}(\gamma^{*}) = \operatorname{supp}(\mathbb P)\times\operatorname{supp}(\mathbb Q)$.
\end{itemize}
\end{corollary}

\paragraph{Remark:} Property (i) allows fast optimization \citep{cuturi2013sinkhorn,genevay2016stochastic}; Properties (ii) and (iii) are essential to a well-defined Y$\mid$X shift we will propose. Compared with other distribution distances, (entropic) optimal transport has a well-established geometric meaning \citep{gangbo1996geometry}, and we will consistently use it to measure distribution shift.

\section{Theoretical Results}
\label{sec:theory}

We assume that covariate $X$ and label $Y$ are distributed on the metric spaces $(\mathcal X,\rho_{\mathcal X})$ and $(\mathcal Y,\rho_{\mathcal Y})$, respectively, and give the rigorous definitions of their domain distribution shifts and essential theorems below.

\subsection{Rigorous Definition of X and Y$\mid$X Shifts}

\begin{definition}[X Shift]
\label{def:x-shift}
Using entropic optimal transport, the X (covariate) shift is defined as:
\begin{align}
S_{Cov}
&=
W_{\beta}\!\bigl(\mathcal D_{X}^{S},\mathcal D_{X}^{T}\bigr)
=
\inf_{\gamma\in\Gamma(\mathcal D_{X}^{S},\mathcal D_{X}^{T})}
\Bigl\{
  \int \rho_{\mathcal X}(x_S,x_T)\, \notag \\[4pt]
&\qquad
  d\gamma(x_S,x_T)
  + \beta\,H\bigl(\gamma \bigm|
     \mathcal D_{X}^{S}\otimes\mathcal D_{X}^{T}\bigr)
\Bigr\} \label{eq:cov-shift}
\end{align}
where the optimal transport plan is denoted as $\gamma^{*}$.
\end{definition}

Note that $\gamma^{*}$ is a joint distribution of $\mathcal D_{X}^{S}$ and $\mathcal D_{X}^{T}$ that gives the minimum transport cost. When $\mathcal D_{X}^{S} = \mathcal D_{X}^{T}$ and $\beta=0$, $\gamma^{*}$ collapses to $
d\gamma^{*}(x_S, x_T) = \mathbf{1}_{\{\,x_S = x_T\,\}}\,d\mathcal D_{X}^{S}(x_S)
$, and $\mathbf{1}_{\{\cdot\}}$ is the indicator function, then $S_{Cov} = W_{1}(\mathcal D_{X}^{S}, \mathcal D_{X}^{S}) = 0$. By Corollary~\ref{cor:entropic-properties}(ii), $\gamma^{*}$ is unique for $\beta>0$.

Also note that we always discuss the Y$\mid$X shift under the realistic stochastic labeling setting, where the label follows a conditional distribution given the covariate. To clarify the shift concept, we introduce four refined definitions. Definitions~\ref{def:point-yx-shift} and \ref{def:total-point-yx-shift} extend the existing Y$\mid$X shift concept to general label space and stochastic labeling, but they remain ill-defined under support mismatch. In contrast, our new shifts given in Definitions~\ref{def:pair-yx-shift} and \ref{def:total-pair-yx-shift} stay well-defined even when supports differ.

\begin{definition}[Point Y$\mid$X Shift]
\label{def:point-yx-shift}
For any $x\in\operatorname{supp}(\mathcal D_{X}^{S})\cap \operatorname{supp}(\mathcal D_{X}^{T})$, the point Y$\mid$X shift is defined as:

\begin{align}
S_{point}(x)
&=
W_{1}\!\bigl(\mathcal D_{Y|X=x}^{S},\mathcal D_{Y|X=x}^{T}\bigr)
=
\inf_{\gamma\in\Gamma(\mathcal D_{Y|X=x}^{S},\mathcal D_{Y|X=x}^{T})} \notag \\[3pt]
&\quad
\int \rho_{\mathcal Y}(y_S,y_T)\,d\gamma(y_S,y_T)
\label{eq:point-shift}
\end{align}

$S_{point}(x)$ measures the Y$\mid$X shift at a single point $x$.
\end{definition}

\begin{definition}[Total Point Y$\mid$X Shift]
\label{def:total-point-yx-shift}
If $\operatorname{supp}(\mathcal D_{X}^{S}) = \operatorname{supp}(\mathcal D_{X}^{T})$, the total point Y$\mid$X shift is the expectation of $S_{point}(x)$ under $\mathcal D_{X}^{S}$ or $\mathcal D_{X}^{T}$, respectively:
\begin{align}
S_{Cpt}^{S} 
&= \mathbb E_{x \sim \mathcal D_{X}^{S}}\bigl[S_{point}(x)\bigr] \notag \\[2pt]
&= \int W_{1}\bigl(\mathcal D_{Y|X=x}^{S}, \mathcal D_{Y|X=x}^{T}\bigr)\,
      d\mathcal D_{X}^{S}(x) \label{eq:cpt-s} \\[6pt]
S_{Cpt}^{T}
&= \mathbb E_{x \sim \mathcal D_{X}^{T}}\bigl[S_{point}(x)\bigr] \notag \\[2pt]
&= \int W_{1}\bigl(\mathcal D_{Y|X=x}^{S}, \mathcal D_{Y|X=x}^{T}\bigr)\,
      d\mathcal D_{X}^{T}(x) \label{eq:cpt-t}
\end{align}

\end{definition}

This definition relies on the support overlap condition $\operatorname{supp}(\mathcal D_{X}^{S}) = \operatorname{supp}(\mathcal D_{X}^{T})$, which avoids ill-definition, although not necessarily true when X shifts. As one can see, the total point Y$\mid$X shift is a generalized form of the existing bound's Y$\mid$X shift:

\begin{corollary}[Relationship to Theorem~\ref{thm:existing-bound}]
\label{cor:relationship-existing-bound}
Assume deterministic labeling $\mathcal D_{Y|X=x}^{S} = \delta_{f_S(x)}$, $\mathcal D_{Y|X=x}^{T} = \delta_{f_T(x)}$, and $\mathcal Y\subset\mathbb R$ with $\rho_{\mathcal Y} = |\cdot|$, then:
\begin{align}
S_{Cpt}^{S} &= \mathbb E_{x \sim \mathcal D_{X}^{S}}\bigl[\,|f_S(x) - f_T(x)|\bigr] \\
S_{Cpt}^{T} &= \mathbb E_{x \sim \mathcal D_{X}^{T}}\bigl[\,|f_S(x) - f_T(x)|\bigr]
\end{align}
which are exactly the Y$\mid$X shift terms in Eq.~\eqref{eq:existing-bound}.
\end{corollary}

To accommodate support mismatch, we suggest comparing Y$\mid$X conditional distributions by point pair $(x_S, x_T)$. We define:

\begin{definition}[Pair Y$\mid$X Shift]
\label{def:pair-yx-shift}
For $x_S \sim \mathcal D_{X}^{S}$ and $x_T \sim \mathcal D_{X}^{T}$, let the conditional label distributions be $\mathcal D_{Y|X=x_S}^{S}$ and $\mathcal D_{Y|X=x_T}^{T}$. The Y$\mid$X shift at point pair $(x_S, x_T)$ is defined as:
\begin{equation}
S_{pair}(x_S, x_T)
= W_{1}\bigl(\mathcal D_{Y|X=x_S}^{S}, \mathcal D_{Y|X=x_T}^{T}\bigr).
\end{equation}
\end{definition}

This definition relaxes the point Y$\mid$X shift in Definition~\ref{def:point-yx-shift} by allowing comparison between any point pairs, making it suitable for $\operatorname{supp}(\mathcal D_{X}^{S}) \neq \operatorname{supp}(\mathcal D_{X}^{T})$.

\begin{definition}[Total Pair Y$\mid$X Shift]
\label{def:total-pair-yx-shift}
Let $\gamma^{*}$ be the optimal transport plan as in Definition~\ref{def:x-shift}---the joint distribution coupling $\mathcal D_{X}^{S}$ and $\mathcal D_{X}^{T}$, then the total pair Y$\mid$X shift is defined as the expectation of $S_{pair}(x_S, x_T)$ under $\gamma^{*}$:
\begin{align}
S_{Cpt}^{\gamma^{*}}
&=
\mathbb E_{(x_S, x_T)\sim \gamma^{*}}\bigl[S_{pair}(x_S, x_T)\bigr] \notag \\[3pt]
&=
\int W_{1}\bigl(\mathcal D_{Y|X=x_S}^{S}, \mathcal D_{Y|X=x_T}^{T}\bigr)\,
   d\gamma^{*}(x_S, x_T)
\label{eq:cpt-gamma}
\end{align}
\end{definition}

This definition is based on the optimal transport plan $\gamma^{*}$ for X shift, which minimizes the total transport cost between $\mathcal D_{X}^{S}$ and $\mathcal D_{X}^{T}$. Intuitively, since $\gamma^{*}$ places most of its mass on nearby pairs $(x_S, x_T)$, $S_{Cpt}^{\gamma^{*}}$ is evaluated mainly on such neighboring points.

\begin{corollary}[Relationship to Definition~\ref{def:total-point-yx-shift}]
\label{cor:total-pair-reduces-to-total-point}
If $\mathcal D_{X}^{S} = \mathcal D_{X}^{T}$ and $\beta=0$, then the plan $\gamma^{*}$ collapses and the total pair Y$\mid$X shift equals to the total point Y$\mid$X shift:
\begin{align}
S_{Cpt}^{\gamma^{*}}
&=
\int W_{1}\bigl(\mathcal D_{Y|X=x_S}^{S}, \mathcal D_{Y|X=x_S}^{T}\bigr)\,
   d\mathcal D_{X}^{S}(x_S) \notag \\[3pt]
&= S_{Cpt}^{S} = S_{Cpt}^{T}
\label{eq:cpt-equality}
\end{align}

\end{corollary}

\begin{theorem}[Uniqueness of Total Pair Y$\mid$X Shift]
\label{thm:uniqueness-total-pair}
Suppose $\mathcal D_{Y|X=x}^{S}$ is unique $\mathcal D_{X}^{S}$-almost everywhere, and $\mathcal D_{Y|X=x}^{T}$ is unique $\mathcal D_{X}^{T}$-almost everywhere. For $\beta>0$, the total pair Y$\mid$X shift $S_{Cpt}^{\gamma^{*}}$ in Definition~\ref{def:total-pair-yx-shift} is unique even when $\operatorname{supp}(\mathcal D_{X}^{S}) \neq \operatorname{supp}(\mathcal D_{X}^{T})$.
\end{theorem}

\paragraph{Remark:} Depending on the entropic optimal transport property (ii) and (iii) in Corollary \ref{cor:entropic-properties}, our total pair Y$\mid$X shift avoids the ill-definition encountered by existing theory and is estimable.

\subsection{General Learning Bound}

We now give a new cross-domain learning error bound based on the total pair Y$\mid$X shift. To be general, the output space of the learner is a metric space $(\mathcal Y^{\prime}, \rho_{\mathcal Y}^{\prime})$, possibly different from the true label space $(\mathcal Y, \rho_{\mathcal Y})$. For the loss function $\ell:\mathcal Y\times\mathcal Y^{\prime}\to\mathbb R$, we require the following basic property from them:

\begin{definition}[Separately Lipschitz Continuity]
\label{def:separately-lipschitz}
For metric spaces $(\mathcal Y, \rho_{\mathcal Y})$ and $(\mathcal Y^{\prime}, \rho_{\mathcal Y}^{\prime})$, a function $\ell:\mathcal Y\times\mathcal Y^{\prime}\to\mathbb R$ satisfies separately $(L_{\ell}, L_{\ell}^{\prime})$-Lipschitz continuity if there exist $L_{\ell}, L_{\ell}^{\prime}\ge0$ such that for any $y_{1}, y_{2}\in\mathcal Y$ and $y_{1}^{\prime}, y_{2}^{\prime}\in\mathcal Y^{\prime}$, there is:
\begin{equation}
\bigl|\ell(y_{1}, y_{1}^{\prime}) - \ell(y_{2}, y_{2}^{\prime})\bigr|
\le L_{\ell}\,\rho_{\mathcal Y}(y_{1}, y_{2})
\;+\; L_{\ell}^{\prime}\,\rho_{\mathcal Y}^{\prime}(y_{1}^{\prime}, y_{2}^{\prime})
\end{equation}
\end{definition}

\paragraph{Remark:} This is a mild assumption: most loss functions are differentiable, which already implies continuity. And their stable optimization usually needs bounded gradients in a region, further ensuring Lipschitz continuity \citep{boyd2004convex}. Note that this assumption also covers asymmetric losses. For instance, for binary classification with label space $[0,1]$, output space is typically $[a,1-a]$ ($a\in(0,0.5)$) by Sigmoid function and $\rho_{\mathcal Y}=\rho_{\mathcal Y}^{\prime}=|\cdot|$, the cross-entropy loss: $
\ell_{CE}(y,\hat y) = -\,y\log\hat y \;-\; (1-y)\log(1-\hat y)
$ satisfies separately $(L_{\ell},L_{\ell}^{\prime})$-Lipschitz continuity with $L_{\ell} = \log\bigl(\tfrac{1-a}{a}\bigr)$ and $L_{\ell}^{\prime} = \tfrac1a$. 

On this basis, we obtain the general cross-domain error bound as follows:

\begin{theorem}[General Learning Bound]
\label{thm:learning-bound}
Given the covariate space $(\mathcal X, \rho_{\mathcal X})$, the label space $(\mathcal Y, \rho_{\mathcal Y})$, and the output space $(\mathcal Y^{\prime}, \rho_{\mathcal Y}^{\prime})$, the source and target distributions are $(\mathcal D_{X}^{S}, \mathcal D_{Y|X=x}^{S})$ and $(\mathcal D_{X}^{T}, \mathcal D_{Y|X=x}^{T})$, respectively. If the loss $\ell:\mathcal Y\times\mathcal Y^{\prime}\to\mathbb R$ satisfies separately $(L_{\ell},L_{\ell}^{\prime})$-Lipschitz continuity, then for any hypothesis $h:\mathcal X\to\mathcal Y^{\prime}$ that satisfies $L_{h}$-Lipschitz continuity, the following bound holds:
\begin{equation}
\epsilon_{T}(h)
\;\le\;
\epsilon_{S}(h)
\;+\;
L_{h}\,L_{\ell}^{\prime}\,S_{Cov}
\;+\;
L_{\ell}\,S_{Cpt}^{\gamma^{*}}
\end{equation}
where $S_{Cov}$ is the X shift in Definition~\ref{def:x-shift} and $S_{Cpt}^{\gamma^{*}}$ is the total pair Y$\mid$X shift in Definition~\ref{def:total-pair-yx-shift}.
\end{theorem}

\paragraph{Remark:} This elegant bound unifies the covariate shift $S_{Cov}$ and the concept shift (total pair Y$\mid$X shift) $S_{Cpt}^{\gamma^{*}}$'s effect on target error via the Lipschitz factors $L_{h}\,L_{\ell}^{\prime}$ and $L_{\ell}$. Notably, it depends only on the Lipschitz continuity of the hypothesis $h$ and the loss $\ell$, without any other specific restriction on the label space $\mathcal Y$ or loss $\ell$. It naturally covers binary classification or regression tasks when $\mathcal Y$ is one-dimensional, and multi-class classification or multi-label tasks when $\mathcal Y$ is multi-dimensional. Besides, since the bound holds under stochastic labeling, it applies to a wide range of supervised learning scenarios in practice. On the other hand, by using the well-defined total pair Y$\mid$X shift, our bound can be tighter than the existing bound in Theorem~\ref{thm:existing-bound}, and more crucially, both our covariate and concept shifts can be robustly estimated from data. Combined with methods for solving the Lipschitz constant $L_{h}$ of learners, our bound is among the first to make rigorous quantification possible for analyzing how domain shifts affect model learning performance in general.

\begin{figure*}[!t]
  \centering
  \subfigure[]{%
    \includegraphics[height=0.27\textwidth]{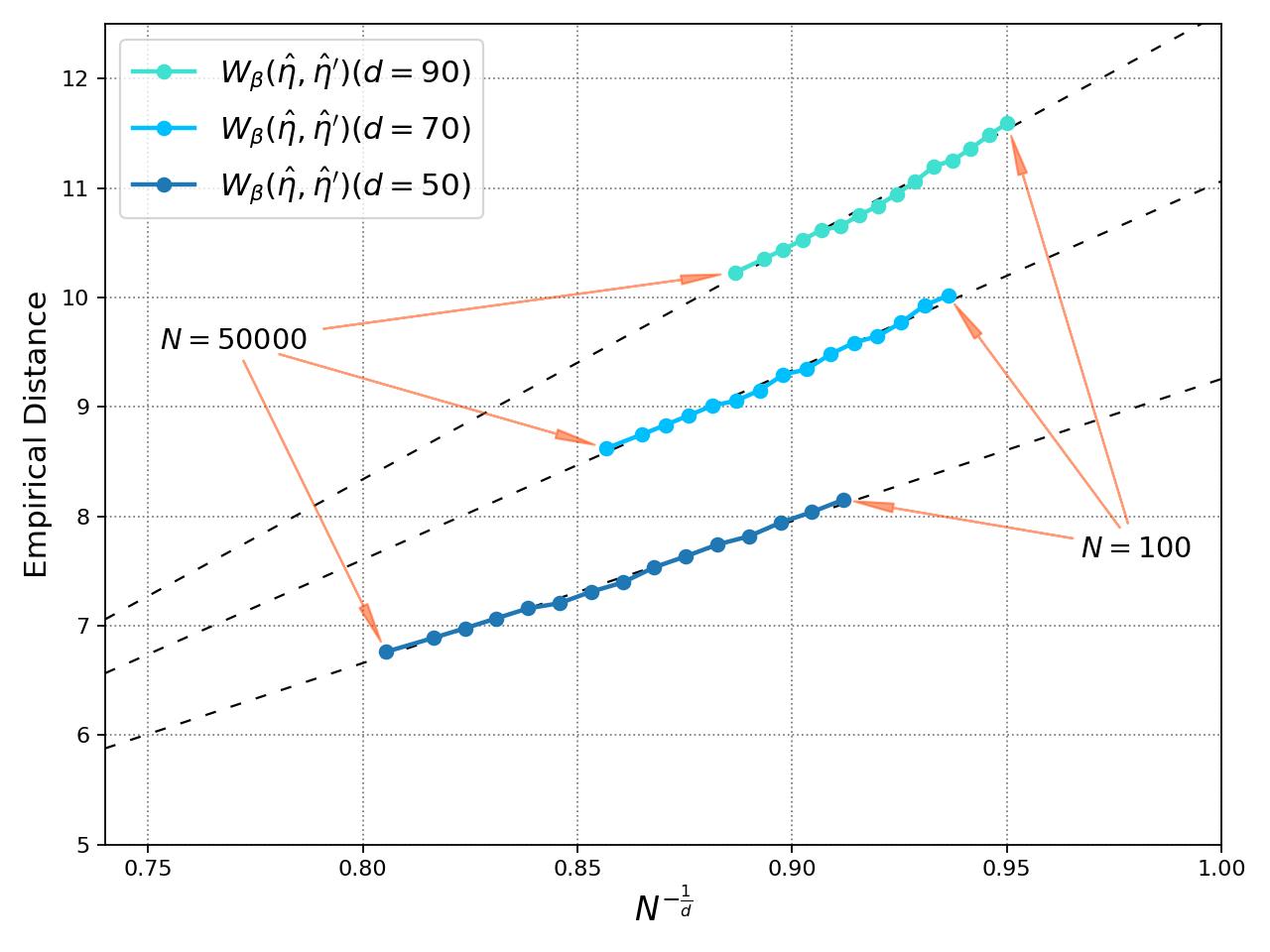}%
    \label{fig:1a}%
  }\hfill
  \subfigure[]{%
    \includegraphics[height=0.27\textwidth]{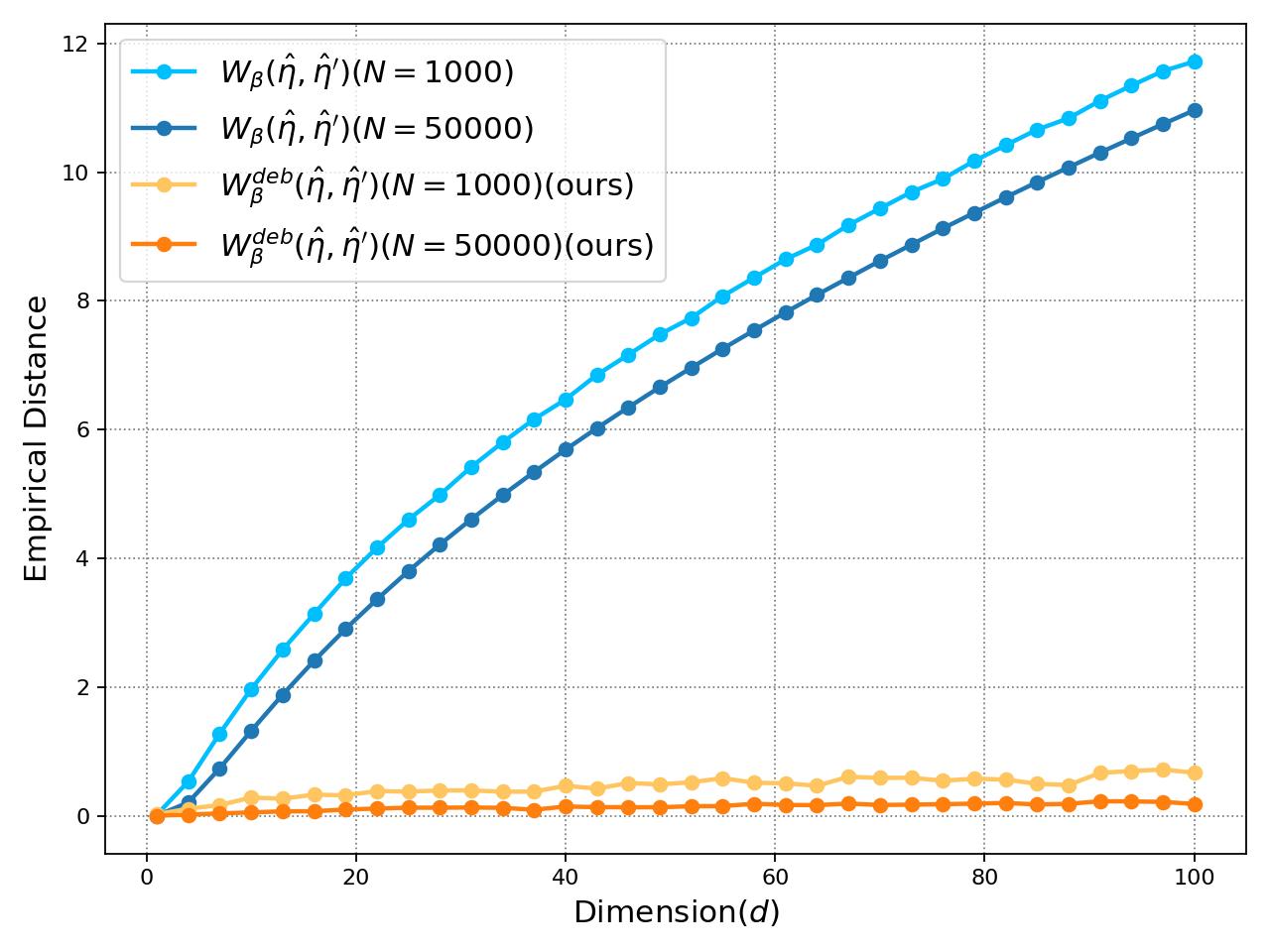}%
    \label{fig:1b}%
  }\hfill
  \subfigure[]{%
    \includegraphics[height=0.27\textwidth]{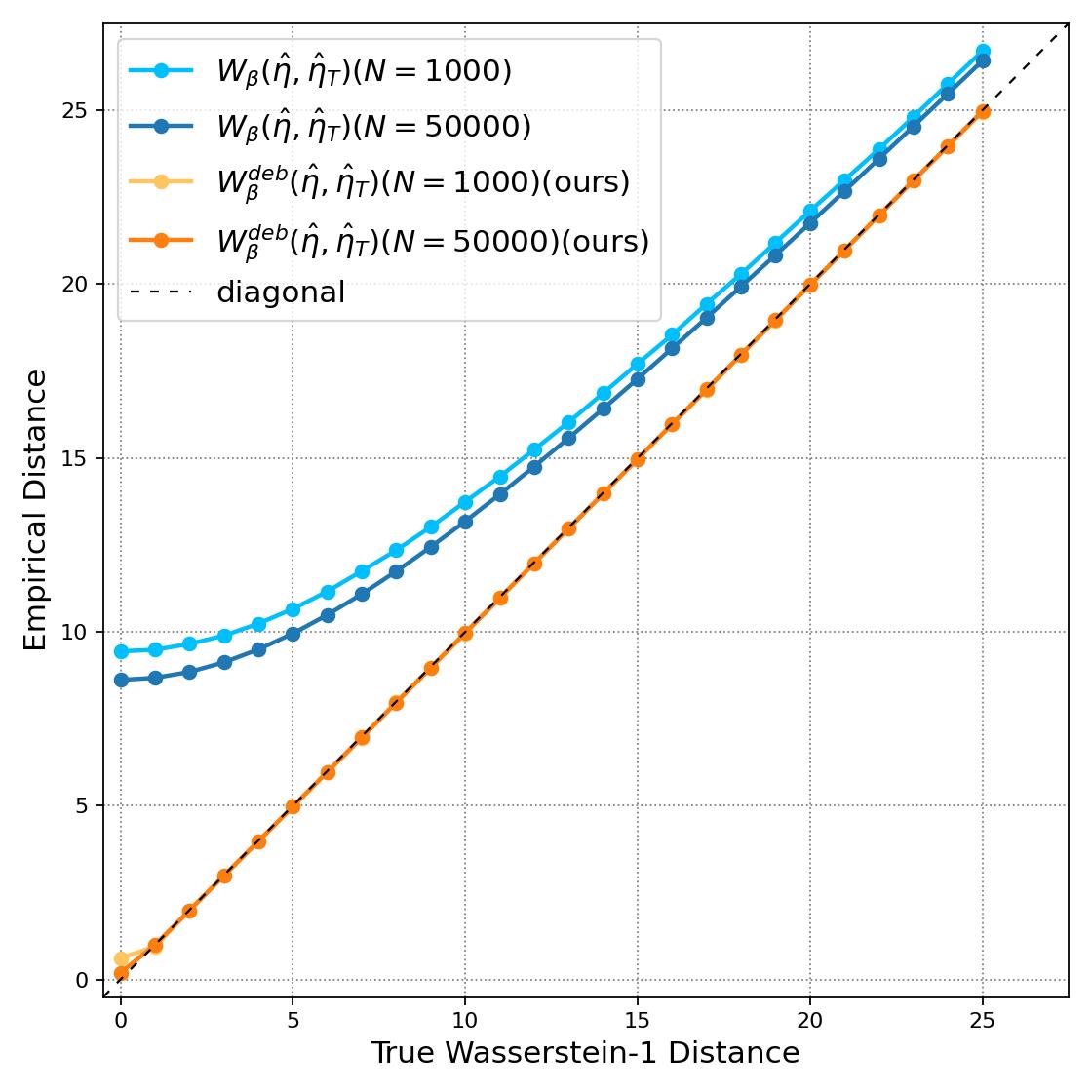}%
    \label{fig:1c}%
  }
  \caption{(a)–(c) show the empirical distance from entropic optimal transport $(\beta = 0.001)$ versus sample size $N$, dimension $d$, and true Wasserstein-1 distance. In (a)–(b), $\hat\eta$ and $\hat\eta'$ are independent empirical measures from high-dimensional standard normals, thus the true distance is zero. The traditional estimator $W_\beta(\hat\eta,\hat\eta')$ greatly overestimates as $d$ increases, as shown in (b), and even a much larger $N$ only brings a small improvement, as shown in (a). In (c), with $d = 70$, $\hat\eta_T$ is the empirical measure of normal $\mathcal{N}(T,I)$, whose true Wasserstein-1 distance to the standard normal is $\|T\|$. Our debiased estimator remains accurate in every case.}
  \label{fig:combined}
\end{figure*}

\section{Estimation of the Learning Bound}
\label{sec:estimation}

In practice, true domain distributions are all unknown. We wish to estimate the shifts from domain samples and analyze how these shifts will influence model performance. Suppose we have i.i.d.\ samples $\{(X_{i}^{(S)}, Y_{i}^{(S)})\}$ and $\{(X_{j}^{(T)}, Y_{j}^{(T)})\}$ drawn from $\mathcal D_{XY}^{S}$ and $\mathcal D_{XY}^{T}$ with sizes $N_S$ and $N_T$, respectively. Estimating our error bound involves three parts: the estimation of X shift, the estimation of Y$\mid$X shift, and the DataShifts algorithm to estimate the overall bound.

\subsection{Estimation of X Shift}

By Definition~\ref{def:x-shift}, the X shift is the entropic optimal transport between $\mathcal D_{X}^{S}$ and $\mathcal D_{X}^{T}$. There already is a traditional estimation method using the entropic optimal transport of empirical distributions—known as the plug-in estimator.

\begin{definition}[Traditional Plug-in Estimator]
\label{def:plugin-estimator}
Given i.i.d.\ samples $\{X_{i}^{(S)}\}\sim\mathcal D_{X}^{S}$, $\{X_{j}^{(T)}\}\sim\mathcal D_{X}^{T}$ with sample sizes $N_S, N_T$ respectively, set the empirical measures: $\widehat{\mathcal D_{X}^{S}} = \frac{1}{N_S}\sum_{i=1}^{N_S} \delta_{X_{i}^{(S)}}$, $\widehat{\mathcal D_{X}^{T}} = \frac{1}{N_T}\sum_{j=1}^{N_T} \delta_{X_{j}^{(T)}}$, their entropic optimal transport is:
\begin{align}
W_{\beta}\bigl(\widehat{\mathcal D_{X}^{S}}, \widehat{\mathcal D_{X}^{T}}\bigr)
&=
\min_{\hat\gamma}\,
  \langle C, \hat\gamma\rangle
  + \beta \sum_{i=1}^{N_S}\sum_{j=1}^{N_T}
    \hat\gamma_{ij}\,\log \hat\gamma_{ij} \notag \\[3pt]
&\quad
+ \beta\,\log\bigl(N_S N_T\bigr) \notag \\[3pt]
&\quad
\text{s.t.}\quad
\hat\gamma \mathbf 1 = \tfrac{1}{N_S}\mathbf 1,\;
\hat\gamma^{\top}\mathbf 1 = \tfrac{1}{N_T}\mathbf 1
\label{eq:empirical-entropic-ot}
\end{align}

where $C\in\mathbb R_{+}^{N_S\times N_T}$ is the cost matrix with $c_{ij} = \rho_{\mathcal X}\bigl(X_{i}^{(S)}, X_{j}^{(T)}\bigr)$, and $\hat\gamma\in \mathbb R_{+}^{N_S\times N_T}$ represents any discretized transport plan satisfying the given linear constraints.
\end{definition}

Such an optimization problem can be solved efficiently at a large scale by the \emph{Sinkhorn} algorithm \citep{cuturi2013sinkhorn}.

\paragraph{Curse of Dimensionality.}
However, in modern applications, the covariate space $\mathcal{X}$ is often high-dimensional. Even when two distributions are very close, their samples' distance can be large. Such curse of dimensionality makes the plug-in estimator greatly overestimated \citep{verleysen2005curse,panaretos2019statistical} (Fig.\ref{fig:1b}). This upward bias decays only in $O(N^{-1/d})$ order \citep{fournier2015rate}, which implies that increasing $N$ has a very limited debiasing effect when $d$ is high (Fig.\ref{fig:1a}). As expected, the bias of the plug-in estimator also decreases as the true distance between distributions grows (Fig.\ref{fig:1c}).

To address the overestimation problem of the traditional plug-in estimator, we propose the following debiased estimator.

\begin{definition}[Debiased Estimator]
\label{def:debiased-estimator}
Given i.i.d.\ samples $\{X_{i}^{(S)}\}\sim\mathcal D_{X}^{S}$, $\{X_{j}^{(T)}\}\sim\mathcal D_{X}^{T}$ with sample sizes $N_S, N_T$ respectively, split the samples in half to obtain four independent empirical measures:
\(
\widehat{\mathcal D_{X}^{S}}' = \frac{2}{N_S} \sum_{i=1}^{N_S/2} \delta_{X_{i}^{(S)}}\),
\(
\widehat{\mathcal D_{X}^{S}}'' = \frac{2}{N_S} \sum_{i=N_S/2+1}^{N_S} \delta_{X_{i}^{(S)}}\),
\(
\widehat{\mathcal D_{X}^{T}}' = \frac{2}{N_T} \sum_{j=1}^{N_T/2} \delta_{X_{j}^{(T)}}\),
\(
\widehat{\mathcal D_{X}^{T}}'' = \frac{2}{N_T} \sum_{j=N_T/2+1}^{N_T} \delta_{X_{j}^{(T)}}
\). The debiased estimator is:
\begin{equation}
\begin{aligned}
W_{\beta}^{deb}\!\bigl(\widehat{\mathcal D_{X}^{S}},\widehat{\mathcal D_{X}^{T}}\bigr)
=
\sqrt{%
\begin{aligned}
\bigl|&
\tfrac12\,W_{\beta}\!\bigl(\widehat{\mathcal D_{X}^{S}}',\widehat{\mathcal D_{X}^{T}}'\bigr)^{2}
+\tfrac12\,W_{\beta}\!\bigl(\widehat{\mathcal D_{X}^{S}}'',\widehat{\mathcal D_{X}^{T}}''\bigr)^{2}
\\[4pt]
&-\tfrac12\,W_{\beta}\!\bigl(\widehat{\mathcal D_{X}^{S}}',\widehat{\mathcal D_{X}^{S}}''\bigr)^{2}
-\tfrac12\,W_{\beta}\!\bigl(\widehat{\mathcal D_{X}^{T}}',\widehat{\mathcal D_{X}^{T}}''\bigr)^{2}
\bigr|
\end{aligned}%
}
\end{aligned}
\label{eq:debias}
\end{equation}

where $W_{\beta}(\hat{\mu}, \hat{\nu})$ is the plug-in estimator of $W_{\beta}(\mu, \nu)$, where $\mu, \nu$ are probability measures.
\end{definition}

\paragraph{Remark:}
The first two terms, \(W_{\beta}\bigl(\widehat{\mathcal D_{X}^{S}}', \widehat{\mathcal D_{X}^{T}}'\bigr)\) and \(W_{\beta}\bigl(\widehat{\mathcal D_{X}^{S}}'', \widehat{\mathcal D_{X}^{T}}''\bigr)\), estimate the distance between $\mathcal D_{X}^{S}$ and $\mathcal D_{X}^{T}$, including the sample bias. And the last two terms,
\(W_{\beta}\bigl(\widehat{\mathcal D_{X}^{S}}', \widehat{\mathcal D_{X}^{S}}''\bigr)\) and \(W_{\beta}\bigl(\widehat{\mathcal D_{X}^{T}}', \widehat{\mathcal D_{X}^{T}}''\bigr)\)
estimate the distance arising from sample bias only. Subtracting the two parts reduces the sample bias, thus giving a better estimate of $W_{\beta}(\mathcal D_{X}^{S}, \mathcal D_{X}^{T})$. Compared with the traditional plug-in estimator, our debiased estimator $W_{\beta}^{deb}$ reduces the overestimation and remains accurate regardless of the true distribution distance (see Fig.\ref{fig:1b} and \ref{fig:1c}).

Moreover, when the covariate space is \emph{Euclidean}: $\mathcal X = \mathbb R^d$, we derived a concentration inequality guaranteeing that the debiased estimator converges to the true distance:

\begin{theorem}[Concentration Inequality of Debiased Estimator]
\label{thm:concentration-debiased}
Let $\mathcal D_{X}^{S}, \mathcal D_{X}^{T}$ be two distributions on $(\mathbb R^{d}, \|\cdot\|)$ with finite squared-exponential moments. For i.i.d.\ samples $\{X_{i}^{(S)}\}\sim\mathcal D_{X}^{S}$, $\{X_{j}^{(T)}\}\sim\mathcal D_{X}^{T}$ with sample sizes $N_S, N_T$ respectively, when $\beta=0$, and for any $\varepsilon>0$, there exists $N$ such that if $N_S, N_T > N$, then:
\begin{align}
\mathbb P\Bigl(
\bigl|\,W_{\beta}^{deb}(\widehat{\mathcal D_{X}^{S}}, \widehat{\mathcal D_{X}^{T}}) 
- W_{\beta}(\mathcal D_{X}^{S}, \mathcal D_{X}^{T})\bigr| 
> \varepsilon
\Bigr)
&\le \notag \\[3pt]
\quad 2\exp\Bigl(-\tfrac{\lambda_S\,N_S\,V_{\varepsilon}\,\varepsilon^{2}}{32}\Bigr)
+ 2\exp\Bigl(-\tfrac{\lambda_T\,N_T\,V_{\varepsilon}\,\varepsilon^{2}}{32}\Bigr) &
\label{eq:deb-bd}
\end{align}

where $\lambda_S, \lambda_T > 0$ depend only on squared-exponential moments of $\mathcal D_{X}^{S}, \mathcal D_{X}^{T}$, respectively, and $V_{\varepsilon} \in [2-\sqrt3, 2)$ depends only on $W_{\beta}(\mathcal D_{X}^{S}, \mathcal D_{X}^{T})/\varepsilon$.
\end{theorem}

\paragraph{Remark:} This theorem implies that the deviation probability of the debiased estimator decays exponentially with sample sizes, so with high probability, our debiased estimator can well approximate the true entropic optimal transport distance $W_{\beta}(\mathcal D_{X}^{S}, \mathcal D_{X}^{T})$. Notably, it also shows the enlightening fact that our estimator's concentration depends not only on each distribution's scale characterized by $\lambda_S, \lambda_T$ but also on the true distance between two distributions.

\subsection{Estimation of Y$\mid$X Shift}

As above, we defined the total pair Y$\mid$X shift $S_{Cpt}^{\gamma^{*}}$ in Definition~\ref{def:total-pair-yx-shift}; we now estimate it from samples.

\begin{definition}[Estimator for Total Pair Y$\mid$X Shift]
\label{def:estimator-total-pair-yx}
Given i.i.d.\ samples $\{(X_{i}^{(S)}, Y_{i}^{(S)})\} \sim \mathcal D_{XY}^{S}$, $\{(X_{j}^{(T)}, Y_{j}^{(T)})\} \sim \mathcal D_{XY}^{T}$ with sample sizes $N_S, N_T$ respectively, the estimator of total pair Y$\mid$X shift is:
\begin{equation}
\hat S_{Cpt}
= \sum_{i=1}^{N_S}\sum_{j=1}^{N_T} 
  \rho_{\mathcal Y}\bigl(Y_{i}^{(S)}, Y_{j}^{(T)}\bigr)\,\hat\gamma_{ij}^{*},
\end{equation}
where $\hat\gamma^{*}\in\mathbb R_{+}^{N_S\times N_T}$ represents the discrete transport plan for $W_{\beta}(\widehat{\mathcal D_{X}^{S}}, \widehat{\mathcal D_{X}^{T}})$ of Definition~\ref{def:plugin-estimator}.
\end{definition}

Note that $\rho_{\mathcal Y}\!\bigl(Y_{i}^{(S)}, Y_{j}^{(T)}\bigr)$ is a single-point estimate of the pair Y$\mid$X shift $S_{pair}\bigl(X_{i}^{(S)}, X_{j}^{(T)}\bigr)$ in Definition~\ref{def:pair-yx-shift}. Such estimation using label pairs may introduce additional looseness in the bound, which we prove can be bounded by the irreducible error \citep{james2013introduction} (also known as the Bayes risk \citep{berger2013statistical}) in traditional statistical learning.

\begin{definition}[Irreducible Error]
\label{def:irreducible-error}
The irreducible error of distribution $\mathcal D_{XY}$ under squared loss $\rho_{\mathcal Y}(\cdot,\cdot)^2$ is defined as:
\begin{equation}
\mathrm I(\mathcal D_{XY})
= \inf_{g:\mathcal X\to\mathcal Y}
  \mathbb E_{(x,y)\sim\mathcal D_{XY}}\bigl[\rho_{\mathcal Y}(y , g(x))^{2}\bigr].
\end{equation}
\end{definition}

The irreducible error is the fundamental error inherent to stochastic labeling and covariate design that no model can overcome. In the deterministic labeling setting with labeling function $f(x)$, $\mathcal D_{Y|X=x} = \delta_{f(x)}$, we have $\mathrm I(\mathcal D_{XY}) = 0$ where $g(x)=f(x)$. In practice, covariate and label are often well correlated, so the irreducible error is small relative to the overall label variability.

Moreover, when label space $\mathcal Y$ is a bounded set in an Euclidean space, we derive a concentration inequality guaranteeing estimator in Definition~\ref{def:estimator-total-pair-yx} approximates true value:

\begin{theorem}[Concentration Inequality for Definition~\ref{def:estimator-total-pair-yx}]
\label{thm:concentration-total-pair}
Let the covariate space $\mathcal X = \mathbb R^d$ with $\mathcal D_{X}^{S}$,$ \mathcal D_{X}^{T}$ having finite squared-exponential moments. Let the label space $\mathcal Y \subset \mathbb R^d$ be bounded by $M = \sup_{y,y'\in\mathcal Y} \|\,y - y'\|$, on which conditional distributions $\mathcal D_{Y|X=x_S}^{S}, \mathcal D_{Y|X=x_T}^{T}$ satisfy $L_{Y|X}$-Lipschitz continuity respectively: $
d_{\mathrm{TV}}\bigl(\mathcal D_{Y|X=x}, \mathcal D_{Y|X=x'}\bigr)
\;\le\; L_{Y|X}\,\|\,x - x'\|$. For i.i.d.\ samples $\{(X_{i}^{(S)}, Y_{i}^{(S)})\} \sim \mathcal D_{XY}^{S}$, $\{(X_{j}^{(T)}, Y_{j}^{(T)})\} \sim \mathcal D_{XY}^{T}$ with sample sizes $N_S, N_T$, when $\beta>0$, and for any $\varepsilon>0$, there exists $N$ such that if $N_S, N_T > N$, then:
\begin{align}
\mathbb P\bigl(\,|\hat S_{Cpt} - S_{Cpt}^{\gamma^{*}} - \varDelta| < \varepsilon\,\bigr)
&\le\; 2\exp\Bigl(-\frac{N_S\,N_T\,\varPhi\,\varepsilon^{2}}{(N_S + N_T)\,M^{2}}\Bigr) \notag \\[4pt]
\quad +\; \exp\Bigl(-\frac{\lambda_S^{1/2}\,N_S\,\varPhi\,\varepsilon^{2}}{4\,\lambda_T^{1/2}\,M^{2}}\Bigr) 
&\quad +\; \exp\Bigl(-\frac{\lambda_T^{1/2}\,N_T\,\varPhi\,\varepsilon^{2}}{4\,\lambda_S^{1/2}\,M^{2}}\Bigr)
\label{eq:cpt-bound}
\end{align}

where $d_{\mathrm{TV}}$ is the total-variation distance, $\lambda_S, \lambda_T > 0$ depend only on the squared-exponential moments of $\mathcal D_{X}^{S}, \mathcal D_{X}^{T}$, $\varPhi > 0$ depends on $L_{Y|X}, \lambda_S, \lambda_T, \beta$, and the constant $\varDelta$ satisfies: $0 \;\le\; \varDelta 
\;\le\; \sqrt{\mathrm I(\mathcal D_{XY}^{S})} \;+\;\sqrt{\mathrm I(\mathcal D_{XY}^{T})}$.

\end{theorem}

\paragraph{Remark:} The term $\varDelta$ is the upward bias due to single-point estimation of $S_{pair}$ and is bounded by the irreducible errors of $\mathcal D_{XY}^{S}$ and $\mathcal D_{XY}^{T}$. In the deterministic labeling, we have $\varDelta = \mathrm I(\mathcal D_{XY}^{S}) = \mathrm I(\mathcal D_{XY}^{T}) = 0$. Since the irreducible errors are usually small, the deviation probability of $\hat S_{Cpt}$ decays exponentially with sample sizes, and our estimator well-approximates the true value with high probability.

\subsection{DataShifts Algorithm}
The Lipschitz constant of the learner have been well studied, such as logistic regression \citep{roux2012stochastic} and neural network \citep{fazlyab2019efficient}. Combined with the Lipschitz constant solver method for the learner, we give the followig \cref{alg:datashifts} (DataShifts) for quantifying X and Y$\mid$X shifts from samples and estimating error bounds.

\begin{algorithm}
   \caption{DataShifts}
   \label{alg:datashifts}
\begin{algorithmic}
   \STATE {\bfseries Input:} samples $\{{(X_{i}^{(S)},Y_{i}^{(S)})}\}$, $\{{(X_{j}^{(T)},Y_{j}^{(T)})}\}$,
   \STATE \quad \quad \quad Lipschitz constant of learner $L_{h}$ (optional),
   \STATE \quad \quad \quad loss function $\ell$ (optional),
   \STATE \quad \quad \quad Source domain empirical error $\hat \epsilon_{S}$ (optional)
   \STATE {\bfseries Do:}
   \STATE  Estimate X shift by \cref{def:debiased-estimator} as $\hat S_{Cov}$
   \STATE Estimate Y$\mid$X shift by \cref{def:estimator-total-pair-yx} as $\hat S_{Cpt}$
   \IF{$\ell$, $L_{h}$, and $\hat \epsilon_{S}$ are provided}
   \STATE  Get separately $(L_{\ell},L_{\ell}^{\prime})$-Lipschitz continuity for $\ell$
   \STATE  Estimate bound: $B=\hat \epsilon_{S}+L_{h}L_{\ell}^{\prime}\hat S_{Cov}+L_{\ell}\hat S_{Cpt}$
   \ENDIF
   \STATE {\bfseries Return:} $\hat S_{Cov}$, $\hat S_{Cpt}$ and $B$ (optional)
\end{algorithmic}
\end{algorithm}

\section{Experiments}
\label{sec:experiments}
\subsection{Novozymes Enzyme Stability Prediction}
The Novozymes Enzyme Prediction Competition \cite{novo2023enzyme} is a large‐scale Kaggle contest. It provided 9 000 point‐mutation samples spanning 180 enzyme families and challenged models to accurately predict the transition temperature of point‐mutated enzymes in a new enzyme family. Each enzyme family is treated as a separate domain; due to distribution shifts between them, thousands of participants found it difficult to develop any effective solution. 

\begin{figure}[ht]
\begin{center}
\includegraphics[width=0.5\textwidth]{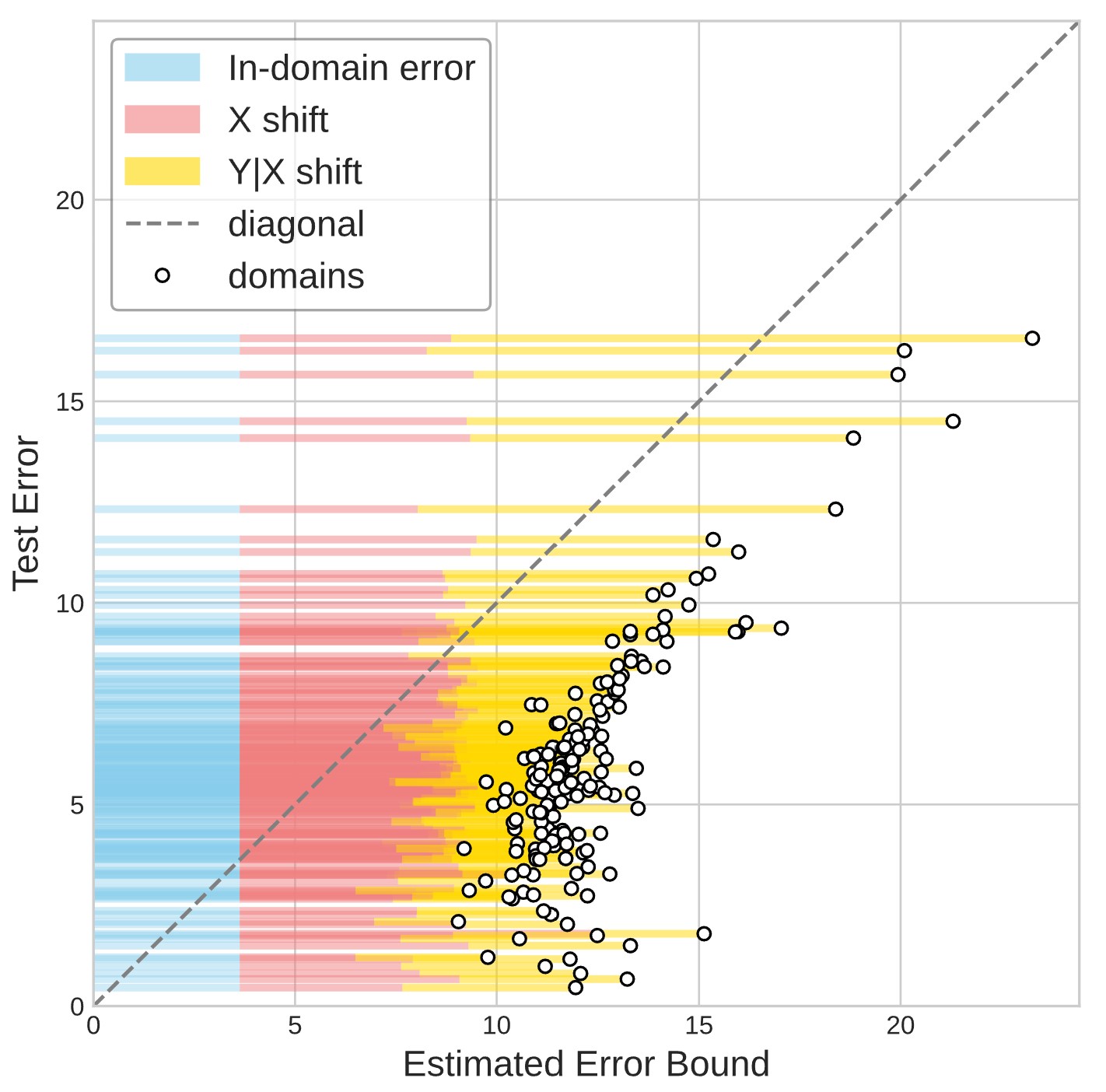}
\caption{Test Error v.s. Estimated Error Bound on 179 Domains}
\label{novo2023fig}
\end{center}
\end{figure}

For this real-world regression task, we selected the enzyme family with the largest number of samples as the source domain and treated the other 179 families as individual target domains. Using the 20 most significant features and the absolute‐error loss which is separately $(1,1)$‐Lipschitz continuous, we trained a 3‐layer MLP on the source domain, then computed both the test error and our error bound by DataShifts algorithm on each of the remaining 179 domains. The results is shown in Fig.\ref{novo2023fig}. 

In this figure, the overall trend of the test error and the error bound lies just below the diagonal, indicating that our tight bound captures the true error across domains effectively. At the same time, it directly shows the contributions of X and Y$\mid$X shifts on the error bound. This results clearly explain the failure of generalization due to distribution shifts.

\section{Conclusion}
\label{sec:conclusion}
We have presented a unified theoretical and algorithmic framework for analyzing generalization under distribution shift. By redefining covariate and concept shift using entropic optimal transport, we derived a new learning bound that is both tighter and estimable from finite samples. Our proposed estimators come with strong concentration guarantees and are integrated into the DataShifts algorithm, enabling accurate pre-deployment error prediction. This work closes a critical gap between theory and practice, offering a rigorous and practical tool for quantifying and analyzing distribution shift in real-world learning.

\section*{Acknowledgements}
Thanks to Jie Ren for suggestions and feedback on this work. This study was funded by the Guangdong Basic and Applied Basic Research Foundation (2024A1515-010699) to LCX.


\bibliography{main}
\bibliographystyle{icml2025}

\end{document}